# Intrinsic-feature-guided 3D Object Detection

Wanjing Zhang[a,b], Chenxing Wang[a,b,*]

[a]*School of Automation, Southeast University, 2# Sipailou, Xuanwu, Nanjing, 210096, China*

[b]*Key laboratory of Measurement and Control of Complex Systems of Engineering, Ministry of Education, Southeast University, Nanjing, 210096, China*

*cxwang@seu.edu.cn

## Abstract

LiDAR-based 3D object detection is essential for autonomous driving systems. However, LiDAR point clouds may appear to have sparsity, uneven distribution, and incomplete structures, significantly limiting the detection performance. In road driving environments, target objects referring to vehicles, pedestrians and cyclists are well-suited for enhancing representation through the complete template guidance, considering their grid and topological structures. Therefore, this paper presents an intrinsic-feature-guided 3D object detection method based on a template-assisted feature enhancement module, which extracts intrinsic features from relatively generalized templates and provides rich structural information for foreground objects. Furthermore, a proposal-level contrastive learning mechanism is designed to enhance the feature differences between foreground and background objects. The proposed modules can act as plug-and-play components and improve the performance of multiple existing methods. Extensive experiments illustrate that the proposed method achieves the highly competitive detection results. Code will be available at https://github.com/zhangwanjingjj/IfgNet.git.

## 1. Introduction

In recent years, 3D object detection has attracted widespread attention as a key technique in autonomous driving systems. Most 3D object detection algorithms are developed based on point cloud data since a point cloud can provide accurate location information and remain unaffected by light and climate. However, there are also some issues for point cloud data: 1) the point clouds are sparse for distant and small objects, resulting in uneven distribution; 2) the foreground point sets are generally incomplete due to self-occlusion or external occlusions. Then the detailed features of target objects cannot be extracted thoroughly, affecting the quality of feature representation and leading to false detections.

To address the issues above, Pyramid-RCNN [1], PDV [2], and IA-SSD [3] are proposed to improve the detection performance for small and distant objects. BtcDet [4], SSN [5], and SIENet [6] solve the occlusion issues by employing point cloud completion techniques. However, completing an object through simple mirroring operations often produces a rough result [4, 5]. Point cloud completion methods based on deep learning can achieve more precise results but increase computational burden [6]. Therefore, we propose a simple and effective method based on prior information to alleviate the abovementioned impact. In autonomous driving systems, the detected objects are usually classified into vehicles, pedestrians, and cyclists. Fortunately, each of these classes displays a rigid common structure. For example, a pedestrian usually comprises a head, body, and limbs, while a cyclist can generally be described as a combination of an upper body and two wheels. Therefore, for each class, we introduce a basic template with a complete shape and dense points to provide rich intrinsic features as prior. Through back-propagation, these prior constraints guide the network to learn more discriminative foreground features.

Since the road environment is usually complex, most false detections arise from background objects resembling the foreground objects. For instance, trees, road signs, and poles are all upright and elongated, which makes them similar to pedestrians and can easily cause interference. Consequently, it is also necessary to suppress the background features to further enhance the feature discrimination between the background and foreground objects. To achieve this, we introduce a contrastive learning method. Since detection tasks aim to achieve object-level differentiation rather than point-level or scene-level differentiation, the proposal-level contrastive learning method is adopted to construct object relations, where labels are assigned to the proposals by calculating Intersection over Union (IoU) between ground-truth (GT) boxes and predictions.

In summary, the main contributions of this study can be concluded as follows:

(1) A template-assisted feature enhancement module is designed to extract intrinsic features as prior for foreground objects. This module improves the network's feature representation capability and addresses issues caused by incomplete structures and uneven distribution of point clouds.

(2) A proposal-level supervised contrastive learning mechanism is constructed to suppress background features and enhance the difference between foreground and background objects.

(3) The two proposed modules can function as plug-and-play components and be integrated into various base detectors.

## 2. Related Work

**2.1 Lidar-based 3D Object Detection Algorithms**

According to the different representations of point clouds, 3D object detection algorithms can be categorized into four types: projection-based methods, voxel-based methods, point-based methods, and voxel-point-based methods.

For projection-based methods, point clouds are projected into the 2D plane, and 2D convolutional neural networks are used to extract features. VeloFCN [7] makes a point cloud projected to a 2D depth map, and then the detection results are obtained through an end-to-end fully convolutional network. Subsequently, PIXOR [8] merges the height, occupancy, and reflectivity information into a single channel to extract the compact representation from the bird-eye-view (BEV) map. RangeRCNN [9], a range-view-based method, employs dilated residual blocks to adapt different object scales, and projects features from the range view to the BEV for detection. These projection-based methods can leverage mature 2D convolutional neural networks for feature extraction, but the projection process inevitably leads to the loss of 3D structural information.

For voxel-based methods, point clouds are divided into equally-sized grids and then 3D convolutional neural networks are used to extract features from the voxel data. VoxelNet [10] is a classic work in this means. However, 3D convolution has a heavy computational complexity. To solve this issue, SECOND [11] introduces 3D sparse convolution to improve efficiency. PointPillars

[12] encodes point clouds into pillars, a specialized voxel type, to create pseudo-images, enabling efficient detection through 2D convolutions. Then, some methods also tried to enhance the voxel features to improve the detection performance. Part-A2 Net [13] designs two modules for accurate proposal detection: a part-aware module developed by U-Net [14] to extract voxel features for estimating foreground points and their intra-object part locations, and a part-aggregation module using RoI-aware point cloud pooling and a series of sparse convolutions to refine the 3D proposals generated through a Region Proposal Network (RPN) head. SA-SSD [15] guides the network to learn structure-aware features by jointly optimizing two auxiliary tasks, foreground segmentation and center estimation. Voxel R-CNN [16] directly integrates multi-scale voxel features into proposals utilizing voxel RoI pooling to achieve both high accuracy and efficiency. However, the voxelization process also leads to the loss of fine-grained structural information.

For point-based methods, raw point clouds are directly utilized for feature learning. Techniques such as PointNet [17] and PointNet++ [18] extract features directly from point clouds. F-PointNet [19] extends PointNet to 3D object detection, which leverages 2D object detectors to generate coarse 2D bounding boxes from images first, then maps them to a 3D frustum space, and finally employs PointNet++ to extract point features from the frustums. This method highly depends on the detection effectiveness of the 2D object detectors. PointRCNN [20] only uses point cloud data for detection, which generates 3D proposals through point cloud segmentation and refines the bounding boxes in the second stage. STD [21] utilizes spherical anchors to reduce computation costs and introduces an IoU branch to enhance localization accuracy. 3DSSD [22] eliminates the up-sampling layer and proposes a fusion sampling strategy based on feature distance to balance the detection efficiency and accuracy. The point-based methods show great advantages in locating objects accurately but remain challenging in computation cost and detection speed.

As analyzed above, the advantages of the voxel and point representations are combined to enhance the detection performance. PV-RCNN [23] employs 3D sparse convolution to extract voxel features and generate 3D proposals. It then samples key points from the original point clouds and utilizes the voxel set abstraction module to derive key point features, which are used to obtain proposal features and refine the bounding boxes.

The above 3D object detection methods perform well but are limited to LiDAR point clouds due to the issues of sparsity, uneven distribution, and incomplete structures. The following sections will introduce the work on addressing these issues.

**2.2 Work on Solving the Issues of Occlusion and Uneven Distribution**

The LiDAR point clouds are usually unevenly distributed and contain some holes due to the occlusion that appears in the scanning process. Pyramid-RCNN [1] proposes some modules to handle the sparsity and uneven distribution of points in proposals, including the RoI-grid pyramid, RoI-grid attention, and density-aware radius prediction. PDV [2] designs a point density-aware voxel network to improve the detection performance for small objects. IA-SSD [3] employs an instance-aware down-sampling strategy to retain more foreground points, addressing the sparsity of point clouds. These methods improve the detection of distant and small objects but still leave the problems of occlusion and incomplete structures in LiDAR data. BtcDet [4] solves the occlusion problem by employing a mirroring operation to restore the target objects with approximately complete shapes and estimating the shape occupancy of the targets. SSN [5] proposes a shape signature network, which employs centro-symmetry for shape completion and then uses the embedded shape vector to guide network learning. SIENet [6] pre-trains the point cloud completion network PCN [24] to generate complete and dense target points, enriching the spatial information. Associate-3Ddet [25] uses objects with high point cloud density as conceptual models and minimizes the feature differences between conceptual models and actual objects to mitigate the occlusion and sparsity issues. However, dense point clouds may still contain holes caused by occlusion.

The above methods complete the point clouds using complex algorithms or even a trained completion network before the true tasks, making the process quite complex. In contrast, we observe that the detected objects in these scenes can be represented by specific templates that provide intrinsic features through their complete shapes and dense point distributions. This approach simplifies the operation and enhances detection efficiency largely.

**2.3 Contrastive Learning for Object Detection**

Contrastive learning divides the sample data into positive and negative sets and then tries to maximize the feature similarity for positive samples and minimize it for negative samples. Contrastive learning also includes supervised methods and self-supervised methods. Due to the lack of labels, self-supervised methods often produce positive samples through data augmentation and randomly select other data to construct negative sample sets. Many studies have demonstrated the effectiveness of self-supervised contrastive learning algorithms for object detection. DenseCL [26] introduces a pixel-level contrastive learning strategy for 2D object detection, which generates positive and negative samples by calculating the similarity between local features. Detco [27] combines the contrastive loss of global images and local patches to enhance image-level and instance-level feature representations. Additionally, InsLoc [28] generates contrastive learning samples by randomly pasting foreground images to different background images, and then the foreground features are aligned by the contrastive loss. For 3D object detection, Point Contrast [29] performs point-level contrastive learning on two transformed views of the given point clouds. STRL [30] and Depth Contrast [31] employ the entire point cloud scene as an instance for contrastive learning. However, the semantic relationships between objects are not involved, which is crucial since distinguishing between objects is necessary for the detection task. Proposal-Contrast [32] treats different views of proposals for objects as positive samples and others as negative samples, then a proposal-level contrastive loss is designed to learn more powerful representations.

However, the feature distance is decreased between samples from the same source but increased between non-homologous samples after self-supervised contrastive learning even if the samples are within the same class. This is disadvantageous to the classification or detection task, which evaluates the feature distance between classes not only samples. To this end, the supervised contrastive learning algorithm [33] is proposed to treat all samples from the same class as positive samples and those from different classes as negative samples, with labeled data. In the ordinary open-source 3D object detection datasets, there have been some labels for GT locations and classes, therefore, we try to assign labels for contrastive learning based on the IoU between GT boxes and proposals. Combined with the intrinsic features enhanced from the proposed templates of target objects, more discriminative features can be learned between different classes, resulting in a more powerful identification performance of object detection.

# 3. Method

The framework of our method is shown in Fig. 1. The Voxel R-CNN is used as the base detector,

where 3D proposals are generated by the RPN, and proposal features are obtained through a voxel RoI pooling module. A template-assisted feature enhancement module extracts the intrinsic features for foreground proposals, guiding the network in learning richer semantic and structural information. A proposal-level supervised contrastive learning mechanism is introduced to further enhance the differences between the positive and negative proposal features. Finally, the bounding boxes are refined based on the optimized proposal features.

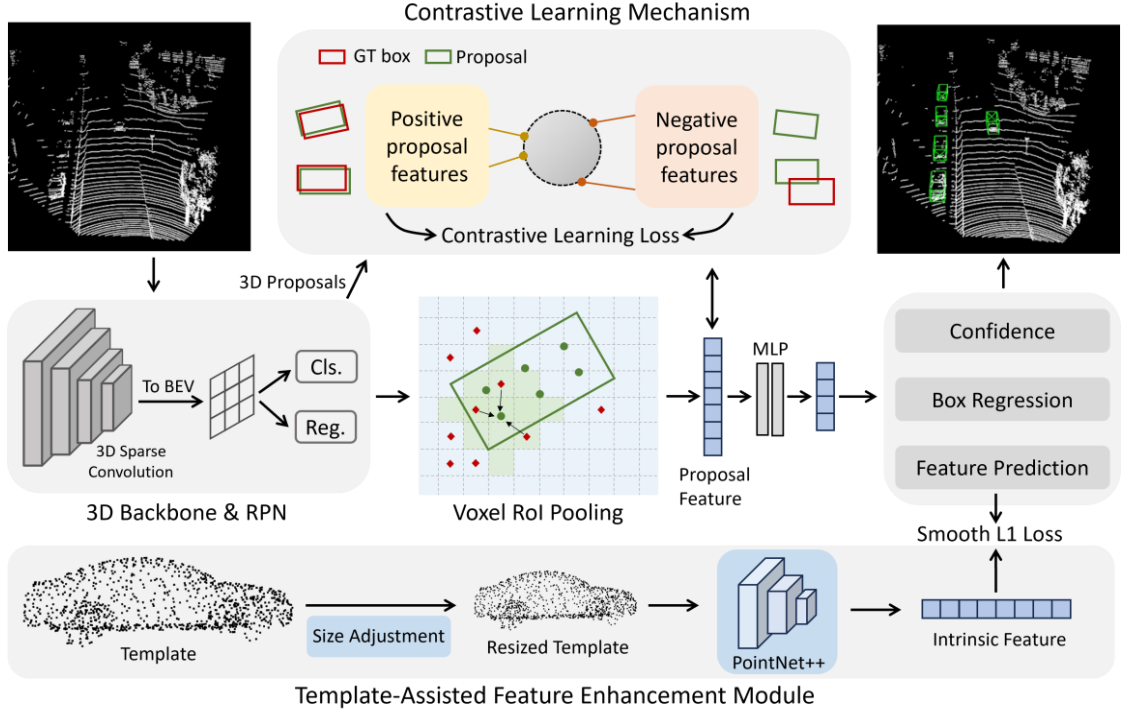

Fig. 1. The framework of our method.

### 3.1 Template-Assisted Feature Enhancement Module

In autonomous driving, 3D object detection based on LiDAR points focuses on the geometry features of the targets, such as vehicles, pedestrians, cyclists, etc. This task becomes difficult if the LiDAR points on these targets are sparse or lack some information due to occlusions. Fortunately, the above targets can be abstracted as some basic geometry information to guide the network to learn the intrinsic features of foreground objects. Therefore, some ordinary templates are designed to provide basic geometry information of targets.

**Template Generation.** The basic geometry information of some targets can be described as follows: a vehicle generally has a body and four wheels, a pedestrian has an upright body and four spindly limbs, and a cyclist consists of an upper human body and two prominent wheels. To be adapted to 3D object detection with point clouds, we present the description of geometry information to some basic template point clouds. These templates are just used to provide the intrinsic features, not to be matched with each instance in detail.

We choose 3D CAD models as the initial templates, and $k$ points are uniformly sampled from each template. The resulting templates are shown in Fig. 2. After obtaining the initial point cloud templates, it is necessary to adjust their sizes and directions according to GT boxes to improve the alignment with actual objects. This adjustment can be described as

$$p'_i = \left(\frac{L}{L'}, \frac{W}{W'}, \frac{H}{H'}\right) \cdot (x_i, y_i, z_i), \tag{1}$$

where $(x_i, y_i, z_i)$ is the original coordinate of a point in template, $(L, W, H)$ and $(L', W', H')$ denote the size of the GT bounding box and template, respectively. Also, the orientation of the coordinate axes is consistent with the local coordinate system of the GT box, making the template direction match that of an actual object.

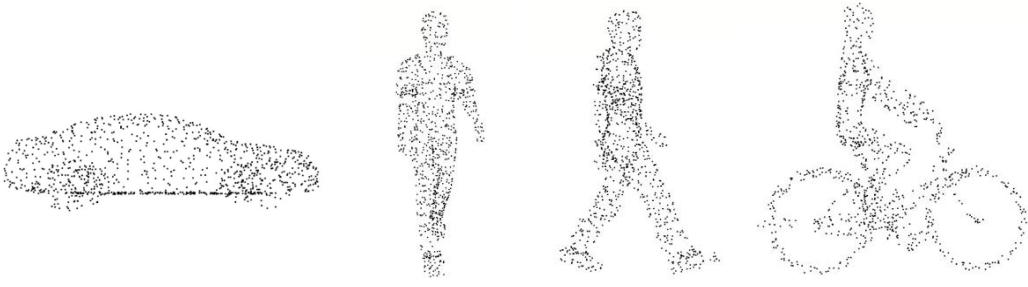

Fig. 2. Illustration of point cloud templates, where the pedestrian template is presented in its front and side views.

**Intrinsic Feature Extraction.** The basic template displays a general structure for each class of objects, so the extracted template features can present the universal and discriminative structural information for a class of objects. Therefore, we apply PointNet++ [18] to extract the template features as intrinsic features to guide network learning. The process of extracting intrinsic features is shown in Fig. 3. The adjusted template point set $\{P_i | i = 1, ..., K\}$ is taken as input, and the farthest point sampling (FPS) algorithm is employed to select $m$ 3D points. Each of the $m$ points is used as a center, and a local region is formed with $k$ neighboring points grouped within a given radius. PointNet [17] is adopted to capture local features from these local regions. To obtain multi-scale context information, we set two different neighboring radii and concatenate the $C$ local features extracted under different receptive fields to form multi-scale features with size $m \times 2C$. Finally, this multi-scale feature is flattened and passed through several fully connected layers to reduce the dimension and obtain the global intrinsic features.

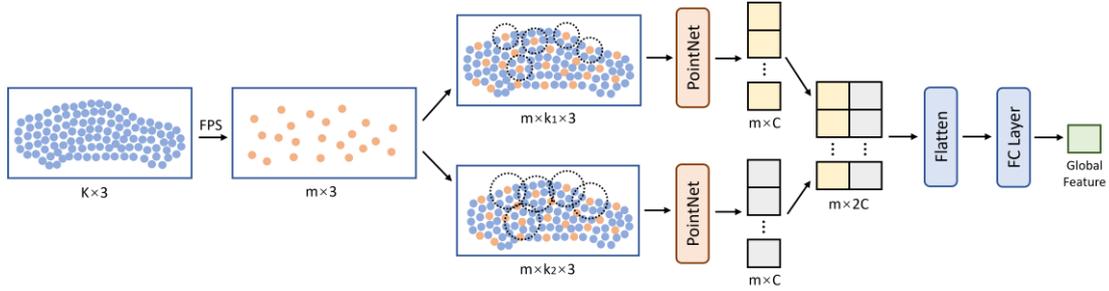

Fig. 3. Illustration of intrinsic feature extraction.

**Enhancement of Foreground Features.** To enhance the foreground proposal features, in the feature prediction module shown in Fig. 1, the proposal feature from the voxel RoI pooling is mapped to a feature vector of the same dimension as the intrinsic feature, by a two-layer MLP. Then, the intrinsic features are used as optimization targets to guide the network in learning more discriminative structural and semantic information for foreground proposals through back-propagation, where the learning process is constrained by Smooth L1 loss.

**3.2 Proposal-Level Supervised Contrastive Learning**

A proposal-level supervised contrastive learning mechanism is introduced to reduce false detections, as shown in Fig. 4. Positive and negative samples are constructed based on the class labels of proposals. The MLP is used to project the proposal features, and the contrastive loss function is designed to guide the learning of positive and negative sample features. This mechanism can extract similar features among the foreground objects of the same class and decrease the feature similarity between foreground and background objects.

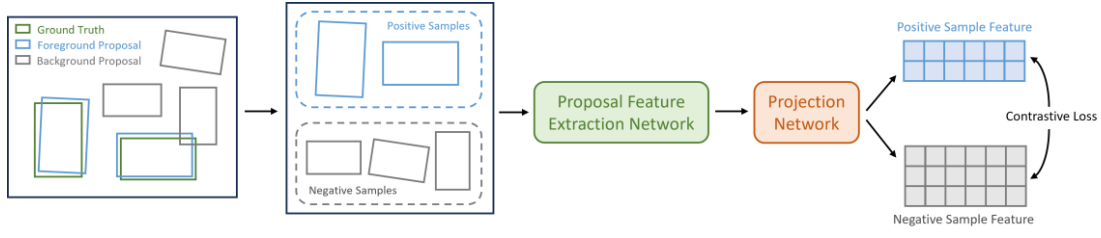

Fig. 4. Illustration of the proposal-level supervised contrastive learning.

**Construction of Positive and Negative Samples.** For a proposal obtained from the RPN, we calculate the IoU between this proposal and all GT boxes and then select the GT box with the highest IoU to be the matched GT box. The IoU threshold for foreground proposals is set as $a$ and for background proposals is set as $b$. The class labels are determined by the IoU between each proposal and its matched GT box, as follows:

1) If the IoU value exceeds $a$, label the proposal as a foreground proposal and assign its class label as a specific number greater than 0;

2) If the IoU value is below $b$, label the proposal as a background proposal and assign its class label as 0.

3) If the IoU value falls between $a$ and $b$, ignore that proposal.

For a given target class, all proposals with this class label form positive samples, while the others with different class labels form negative samples.

**Supervised Contrastive Loss.** A projection network is used to further process the output from the proposal feature extraction network, mapping high-dimensional features into a low-dimensional feature space. Since the projection network is located at a high level within the overall detection network, it can learn more advanced and distinctive feature representations, and these features are more task-relevant and can better support the subsequent detection task. The projection network is instantiated as the MLP, which outputs feature vectors of size $\beta$. Then, these features are utilized to calculate the contrastive loss described as [33]:

$$L_{contra} = \sum_{i \in I} \frac{-1}{|P(i)|} \sum_{p \in P(i)} \log L_i,$$

$$L_i = \frac{\exp\left(f_i \cdot \frac{f_p}{\tau}\right)}{\sum_{a \in A(i)} \exp\left(f_i \cdot \frac{f_a}{\tau}\right)}, \qquad (2)$$

where $I = \{1,2,\ldots,N\}$, $N$ represents the number of proposals in a batch, $A(i) = I/\{i\}$ denotes the indices of proposals excluding the current one, $P(i) = \{p \in A(i): y_p = y_i\}$ represents the index of a proposal that belongs to the same class as the current proposal, $|P(i)|$ is the number of proposals that share the same class as the current proposal, $f$ represents the proposal features output by the projection network, and $\tau \in R^+$ is the temperature coefficient.

As can be seen from the above formula, greater feature similarity within the same class increases the numerator of $L_i$, while smaller similarity between different classes decreases the denominator of $L_i$, together reducing the contrastive loss. Therefore, through gradient back-propagation, the network minimizes the feature distances between proposals of the same class and maximizes the feature distances between those of different classes. For a given class of foreground proposals, contrastive learning enhances the proposal features of that class while suppressing the proposal features of different classes (including the background), which enables the network to distinguish the proposal classes better and reduce false identification.

### 3.3 Framework Constraints

The overall loss consists of two parts: $L_{rpn}$ generated by the RPN, and $L_{rcnn}$ generated by the box refinement network.

**RPN Loss $L_{rpn}$.** For the RPN training, the class labels and regression targets of the anchors are obtained by the IoU-based allocation method. The bounding box is parameterized as $(x, y, z, l, w, h, \theta)$, and the regression targets can be represented as $(t_x^a, t_y^a, t_z^a, t_w^a, t_l^a, t_h^a, t_\theta^a)$, which are calculated as follows:

$$t_x^a = \frac{x_g - x_a}{d_a}, t_y^a = \frac{y_g - y_a}{d_a}, t_z^a = \frac{z_g - z_a}{h_a},$$

$$t_w^a = \log(\frac{w_g}{w_a}), t_l^a = \log(\frac{l_g}{l_a}), t_h^a = \log(\frac{h_g}{h_a}),$$

$$t_\theta^a = \theta_g - \theta_a, \quad (3)$$

where $d_a = \sqrt{(l_a)^2 + (w_a)^2}$ is the diagonal line at the bottom of an anchor.

The RPN loss $L_{rpn}$ is divided into classification loss $L_{cls}$ and bounding box regression loss $L_{reg}$, which can be represented as follows:

$$L_{rpn} = \frac{1}{N_{fg}} \left[ \sum_i L_{cls}(p_i^a, p_i^t) + \mathbb{1}(p_i^t \geq 1) \sum_i L_{reg}(\delta_i^a, \delta_i^t) \right], \quad (4)$$

where $N_{fg}$ represents the number of foreground anchors, $p_i^a$ and $p_i^t$ represent the classification prediction and label for the $i$-th anchor, respectively, and Focal loss [34] is adopted to calculate $L_{cls}$; $p_i^t$ and $\delta_i^t$ represent the regression prediction and target for the $i$-th anchor, respectively, and Smooth L1 loss is used to calculate $L_{reg}$, $\mathbb{1}(p_i^t \geq 1)$ indicates that only foreground anchors participate in calculating $L_{reg}$.

**Box Refinement Loss $L_{rcnn}$.** The loss $L_{rcnn}$ for refining bounding boxes consists of four components: the confidence prediction loss $L_{conf}$, the bounding box regression loss $L_{reg}$, the template-assisted feature optimization loss $L_{temp}$, and the supervised contrastive loss $L_{contra}$. The calculation of $L_{reg}$ is the same as that in the RPN stage, and $L_{contra}$ is given in Sec. 3.2.

The calculation of $L_{conf}$ adopts the Binary Cross Entropy loss. For the $i$-th proposal, its confidence label $y_i$ is determined as follows:

$$y_i = \min(1, \max(0, 2 \times IoU_i - 0.5)), \quad (5)$$

where $IoU_i$ is the IoU between the $i$-th proposal and its corresponding GT boxes.

The calculation of $L_{temp}$ adopts the Smooth L1 loss and takes the intrinsic features obtained in Sec. 3.1 as regression targets:

$$L_{temp} = \frac{1}{N_p} \left[ \mathbb{1}(IoU_i > \mu) \sum_i \text{SmoothL1}(\alpha_i - \alpha_i^t) \right], \quad (6)$$

where $N_p$ represents the number of foreground proposals, $\alpha_i$ and $\alpha_i^t$ represent the proposal feature and the template global feature, respectively, $\mathbb{1}(IoU_i > \mu)$ indicates that only proposals with $IoU_i > \mu$ participate in the calculation. Furthermore, the proposals involved in this loss calculation are consistent with those used in the box regression loss $L_{reg}$.

# 4. Experiments

In this section, we first introduce the datasets in Sec. 4.1. Then, the implementation details are provided in Sec. 4.2. In Sec. 4.3 and 4.4, we validate the effectiveness of our method on the KITTI [35] and Waymo Open [36] datasets, respectively. Finally, we conduct extensive ablation studies to analyze each proposed module in Sec. 4.5.

**4.1 Datasets**

**KITTI Dataset.** The KITTI dataset is widely used for 3D object detection in autonomous driving scenarios. It consists of 7,481 training samples and 7,518 testing samples, and the training samples are further divided into a train split (3,712 samples) and a validation split (3,769 samples). Additionally, the KITTI dataset categorizes objects into three levels: easy, moderate, and hard, according to their size, occlusion, and truncation extents, etc.

**Waymo Open Dataset.** The Waymo Open dataset has more large scale and so provides a more challenging benchmark for 3D object detection. It consists of 798 training sequences (158,361 samples) and 202 validation sequences (40,077 samples). The target objects are categorized into two difficulty levels according to the point number: LEVEL1 (objects with more than 5 points) and LEVEL2 (objects with points ranging from 1 to 5). Additionally, the objects are classified into three classes according to different distance ranges: 0-30m, 30-50m, and 50m-inf.

**4.2 Implementation Details**

**Network Architecture.** For the KITTI dataset, the detection ranges for X, Y, and Z axes are [0, 70.4], [-40, 40], and [-3, 1], respectively (unit in m), which are [-75.2, 75.2], [-75.2, 75.2] and [-2, 4], respectively, for the Waymo dataset. To validate the generalization of our method, we implement our method based on three classic baselines: PointRCNN, Voxel R-CNN, and PV-RCNN, and maintain the same parameter settings of these baselines.

For the proposed template-assisted feature enhancement module, the number of sampling points $k$ for each template is 1024. In the feature extraction process for templates, as shown in Fig. 3, the number of sampled center $m$ is set as 128, and the two neighboring radii are 0.2m and 0.4m. The maximum number of points, $k_1$ and $k_2$, within the two neighborhoods are 16 and 32, respectively. The dimension $C$ of the local feature is 32, and the dimension of the global intrinsic feature is 16.

For the proposal-level supervised contrastive learning module, the IoU threshold $a$ for foreground proposals is set as 0.75, while the IoU threshold $b$ for background proposals is set as 0.25. The output size $\beta$ of the projection network is 128, and the temperature coefficient $\tau$ is 0.1.

**Training and Inference Details.** Our framework is trained in an end-to-end manner with the ADAM optimizer. For the KITTI dataset, we train the network with batch size 8, learning rate 0.01 for 80 epochs with a single A6000 GPU. For the Waymo Open dataset, we train the network with batch size 16, learning rate 0.01 for 30 epochs with two A6000 GPUs. The one-cycle policy is adopted for the learning rate adjustment. During training, non-maximum-suppression (NMS) with an IoU threshold of 0.8 is used to retain 128 proposals generated from the RPN, where the number of positive and negative samples is balanced, and the IoU threshold for positive samples is set to 0.55. During inference, for the KITTI dataset, NMS with an IoU threshold of 0.7 is adopted to keep 100 proposals. For the Waymo Open dataset, 500 proposals are retained using NMS with an IoU threshold of 0.8. Finally, for the refined 3D bounding boxes, NMS with an IoU threshold of 0.1 is applied to remove the redundant boxes.

### 4.3 Evaluation on KITTI Dataset

The detection results are evaluated by the average precision (AP) with an IoU threshold of 0.7 for cars and 0.5 for pedestrians and cyclists. We conduct evaluations on both the validation split and the test set. For the validation split, we directly train the network using the train split and calculate the AP with 11 recall positions. For the test set, the network is trained with 80% of all train+val samples, and the AP are calculated with 40 recall positions.

**Results on Validation Split**

On the validation split, we analyze the performance of our method on three baseline methods: PointRCNN, Voxel R-CNN, and PV-RCNN. The experimental results are shown in Table 1, where the pedestrian and cyclist detection results for PointRCNN and Voxel R-CNN are reproduced based on the publicly available source code.

| Methods | Car AP$_{3D}$ (%) | | | Ped. AP$_{3D}$ (%) | | | Cyc. AP$_{3D}$ (%) | | | FPS (Hz) |
|---|---|---|---|---|---|---|---|---|---|---|
| | Easy | Mod. | Hard | Easy | Mod. | Hard | Easy | Mod. | Hard | |
| PointRCNN [20] | 88.88 | 78.63 | 77.38 | 63.92 | 55.87 | 51.41 | 86.63 | 72.56 | 67.24 | 14.54 |
| **Ours-P** | **89.10** | **79.71** | **78.75** | **64.21** | **59.18** | **52.78** | **87.74** | **73.88** | **69.17** | **14.86** |
| Voxel R-CNN [16] | 89.41 | 84.52 | 78.93 | 66.93 | 60.67 | 55.24 | 86.32 | 73.06 | 69.42 | 27.14 |
| **Ours-V** | **89.85** | **85.18** | **79.48** | **69.67** | **63.18** | **58.63** | **88.80** | **74.52** | **70.97** | **27.32** |
| PV-RCNN [23] | 89.35 | 83.69 | 78.70 | 64.60 | 57.90 | 53.23 | 85.22 | 70.47 | 65.75 | 12.23 |
| **Ours-PV** | **89.47** | **84.10** | **79.15** | **67.85** | **61.28** | **56.37** | **87.23** | **72.25** | **67.20** | **12.57** |

Table 1. 3D detection results on the KITTI validation split with AP calculated by 11 recall positions.

Our method achieves higher performance than the baselines across all levels. Specifically, compared to the baselines (PointRCNN, Voxel R-CNN, PV-RCNN), our method improves the 3D AP for the moderate car class by 1.08%, 0.66%, and 0.41% respectively. For the moderate pedestrian class, our method improves the 3D AP by 3.31%, 2.51%, and 3.38%, and for the moderate cyclist class, our method improves the 3D AP by 1.32%, 1.46%, and 1.78%. It is worth noting that the improvement is relatively small for the car class but more significant for the pedestrian and cyclist classes. This is acceptable because cars usually have more distinct geometric shapes and relatively denser points, providing more information for 3D detection. Therefore, the baseline methods can already perform well in the car class. However, pedestrians and cyclists have small sizes and sparse point clouds, which makes the detection task more challenging. The existing baseline methods have difficulty in capturing the features of small objects, and our method leverages the template-assisted feature enhancement module and contrastive learning mechanism, allowing better learning of the semantic and structural information of pedestrians and cyclists, and resulting in significant performance improvements in these two classes.

In terms of inference speed, both the contrastive learning mechanism and template-assisted feature enhancement module are only introduced during the training stage to learn richer foreground features and suppress background features. In the testing stage, these two modules are removed, so there is no additional computation cost. We measure the running speed on single A6000 GPU for all models, and the results are shown in Table 1. As expected, compared to the baselines, the running speed of our method remains almost unchanged.

The Voxel R-CNN shows the best results through the three baselines, so we just select Voxel R-CNN for comparison, as shown in Fig. 5. Compared to Voxel R-CNN, our method reduces false detection on distant objects significantly, and improves robustness in detecting small objects such as pedestrians.

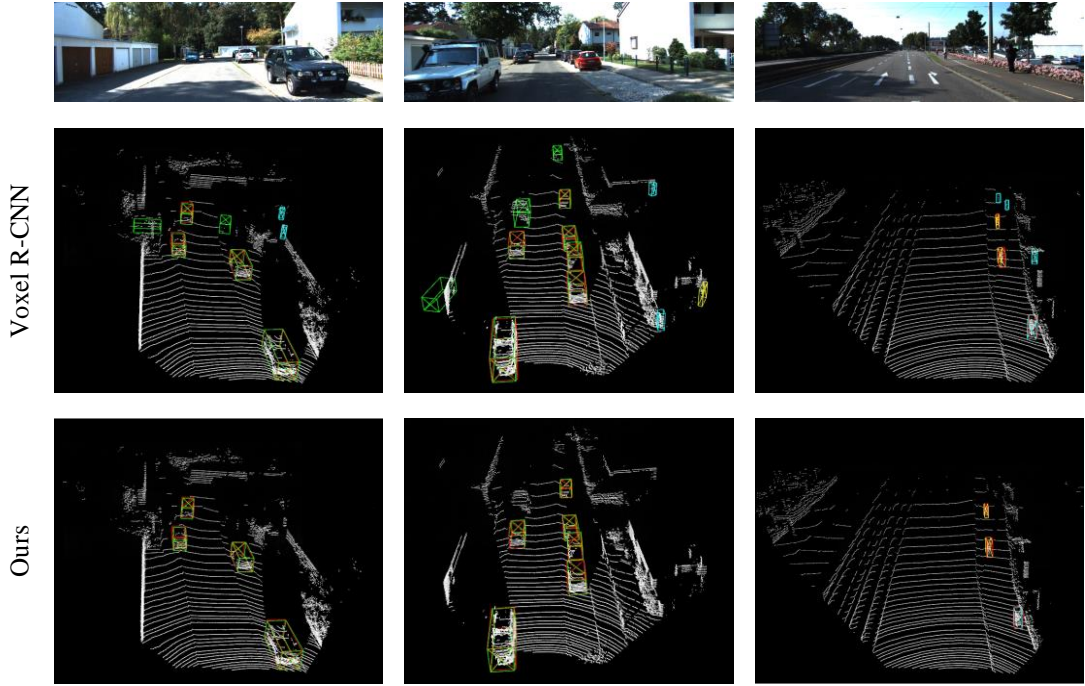

Fig. 5. Qualitative results of Voxel R-CNN and our model on the KITTI validation split. The GT boxes are shown in red, and the predicted bounding boxes for cars, pedestrians, and cyclists are shown in green, blue, and yellow, respectively.

The quantity and structural integrity of point clouds are closely related to the distance of objects and sensors. Therefore, we evaluate our method across different distance ranges to analyze whether it effectively alleviates the problems of sparsity, incomplete structure, and uneven distribution. The KITTI validation split is divided into three categories: 0-20m, 20-40m, and 40m-Inf. We also select Voxel R-CNN as the baseline detector and the experimental results are shown in Table 2. Our method achieves increased AP across all three distance ranges, with greater improvements in the far-distance range compared to the near-distance range, which demonstrates that our method is effective at enhancing the performance of distant objects with fewer points and incomplete structures, thereby mitigating the problem of uneven distribution. In addition, for small objects with sparse point clouds, such as pedestrians, our method outperforms the baseline across all three distance ranges, achieving 3D AP improvements of 1.65%, 2.30%, and 1.86%, respectively.

| Methods | Car (Mod.) (%) | | | Ped. (Mod.) (%) | | | Cyc. (Mod.) (%) | | |
| --- | --- | --- | --- | --- | --- | --- | --- | --- | --- |
| | 0-20m | 20-40m | 40m-Inf | 0-20m | 20-40m | 40m-Inf | 0-20m | 20-40m | 40m-Inf |
| Voxel R-CNN [16] | 90.35 | 82.78 | 45.41 | 71.54 | 41.03 | 5.40 | 96.91 | 66.30 | 46.47 |
| **Ours-V** | **90.46** | **83.81** | **47.03** | **73.19** | **43.33** | **7.26** | **97.57** | **69.19** | **48.98** |
| Increasing value | 0.11 | 1.03 | 1.62 | 1.65 | 2.30 | 1.86 | 0.66 | 2.89 | 2.51 |

Table 2. Performance comparison across different distance ranges on the KITTI validation split, where the moderate AP is calculated by 11 recall positions.

**Results on Test Set**

Table 3 shows the performance of our method on the KITTI test set, using Voxel R-CNN as the base detector. Benefiting from the two proposed modules, our method achieves either the best or highly competitive performance across all difficulty levels in the three target classes. Additionally, we calculate the average AP values for each method across the three difficulty levels. Our method achieves the highest average AP values for all target classes. Compared with the approaches [1-4, 6,

25] mentioned in Sec. 2.2, which similarly aim to address issues of sparsity, uneven distribution and occlusion, our method outperforms all of these approaches in general.

| Methods | Car AP$_{3D}$ (%) | | | | Ped. AP$_{3D}$ (%) | | | | Cyc. AP$_{3D}$ (%) | | | |
|---|---|---|---|---|---|---|---|---|---|---|---|---|
| | Easy | Mod. | Hard | Avg. | Easy | Mod. | Hard | Avg. | Easy | Mod. | Hard | Avg. |
| PointRCNN [20] | 86.96 | 76.64 | 70.70 | 78.10 | 47.98 | 39.37 | 36.01 | 41.12 | 74.96 | 58.82 | 52.53 | 62.10 |
| SA-SSD [15] | 88.75 | 79.79 | 74.16 | 80.90 | - | - | - | - | - | - | - | - |
| Pyramid-PV [1] | 88.39 | 82.08 | 77.49 | 82.65 | - | - | - | - | - | - | - | - |
| PDV [2] | 90.43 | 81.86 | 77.36 | 83.22 | - | - | - | - | **83.04** | 67.81 | 60.43 | 70.43 |
| IA-SSD [3] | 88.34 | 80.13 | 75.04 | 81.17 | 46.51 | 39.03 | 35.60 | 40.38 | 78.35 | 61.94 | 55.70 | 65.33 |
| BtcDet [4] | 90.64 | <u>82.86</u> | **78.09** | <u>83.86</u> | 47.80 | 41.63 | 39.30 | 42.91 | 82.81 | 67.68 | 60.81 | 70.43 |
| SIENet [6] | 88.22 | 81.71 | 77.22 | 82.38 | - | - | - | - | <u>83.00</u> | 67.61 | 60.09 | 70.23 |
| Associate-3Ddet [25] | **91.53** | 80.77 | 75.23 | 82.51 | 45.18 | 37.22 | 34.62 | 39.01 | 72.82 | 57.60 | 51.53 | 60.65 |
| PV-RCNN [23] | 88.57 | 81.49 | 76.75 | 82.87 | <u>48.62</u> | <u>42.03</u> | <u>39.81</u> | <u>43.45</u> | 79.83 | 64.29 | 58.48 | 67.53 |
| PG-RCNN [37] | 89.38 | 82.13 | 77.33 | 82.95 | - | - | - | - | 82.77 | <u>67.82</u> | 61.25 | <u>70.61</u> |
| PVT-SSD [38] | 90.65 | 82.29 | 76.85 | 83.26 | - | - | - | - | - | - | - | - |
| DFAF3D [39] | 88.59 | 79.37 | 72.21 | 80.39 | 47.58 | 40.99 | 37.65 | 42.07 | 82.09 | 65.86 | 59.02 | 68.99 |
| OFFNet [40] | 85.81 | 79.68 | 75.41 | 80.30 | 48.59 | 41.66 | 38.73 | 42.99 | 80.46 | 66.29 | <u>61.33</u> | 69.36 |
| Fuzzy-NMS [41] | 87.30 | 78.71 | 74.03 | 80.01 | 48.38 | 40.01 | 37.31 | 41.90 | 77.53 | 62.05 | 55.18 | 64.92 |
| CFPC [42] | 89.04 | 81.97 | 77.42 | 82.81 | - | - | - | - | - | - | - | - |
| <u>V</u>oxel R-CNN [16] | 90.90 | 81.62 | 77.06 | 83.19 | - | - | - | - | - | - | - | - |
| **Ours-<u>V</u>** | <u>91.26</u> | **82.87** | <u>77.94</u> | **84.02** | **50.48** | **43.08** | **40.96** | **44.84** | 82.33 | **68.05** | **61.52** | **70.63** |

Table 3. 3D detection results on the KITTI test set with AP calculated by 40 recall positions. **Bolded** and <u>underlined</u> values are the best and second-best performance, respectively.

### 4.4 Evaluation on Waymo Open Dataset

To further validate the effectiveness of our method, we conduct experiments on the large-scale Waymo Open dataset. The AP and average precision weighted by heading (APH) are adopted for evaluation. Similar to the KITTI dataset, the IoU threshold is set to 0.7 for vehicles and 0.5 for pedestrians and cyclists.

| Difficulty | Methods | Veh. (%) | | Ped. (%) | | Cyc. (%) | |
|---|---|---|---|---|---|---|---|
| | | mAP | mAPH | mAP | mAPH | mAP | mAPH |
| LEVEL1 | PointRCNN [20] | 43.93 | 43.41 | 18.93 | 16.96 | 45.43 | 43.96 |
| | **Ours-<u>P</u>** | **45.65** | **45.19** | **21.87** | **19.94** | **48.26** | **46.83** |
| | <u>V</u>oxel R-CNN [16] | 75.61 | 74.98 | 74.34 | 63.75 | 69.88 | 68.27 |
| | **Ours-<u>V</u>** | **77.47** | **76.92** | **77.40** | **66.84** | **73.14** | **71.60** |
| | <u>PV</u>-RCNN [23] | 75.65 | 75.00 | 72.78 | 61.35 | 66.80 | 64.89 |
| | **Ours-<u>PV</u>** | **77.28** | **76.68** | **75.13** | **63.64** | **69.94** | **68.08** |
| LEVEL2 | PointRCNN [20] | 37.91 | 37.46 | 15.85 | 14.19 | 43.69 | 42.28 |
| | **Ours-<u>P</u>** | **39.76** | **39.37** | **18.74** | **17.12** | **46.50** | **45.12** |
| | <u>V</u>oxel R-CNN [16] | 66.82 | 66.26 | 65.45 | 55.90 | 67.31 | 65.75 |
| | **Ours-<u>V</u>** | **68.85** | **68.35** | **68.58** | **59.08** | **70.46** | **68.96** |
| | <u>PV</u>-RCNN [23] | 67.08 | 66.48 | 63.86 | 53.63 | 64.34 | 62.51 |
| | **Ours-<u>PV</u>** | **68.70** | **68.17** | **66.24** | **56.05** | **67.30** | **65.66** |

Table 4. Comparisons on the Waymo Open Dataset with 202 validation sequences.

We reproduce the results of the three baseline detectors on the Waymo Open validation set using publicly available source code, and compare them with ours, as shown in Table 4. Our method outperforms the baseline methods across different classes and difficulty levels. On level 1 difficulty, our method improves the APH for the vehicle class by 1.78%, 1.94%, and 1.68%, respectively. For the pedestrian and cyclist classes, the improvements are more significant, with APH increases of 2.98%, 3.09%, and 2.29% for pedestrians, and 2.87%, 3.33%, and 3.19% for cyclists. On level 2 difficulty, our method improves the APH for the vehicle class by 1.91%, 2.09%, and 1.69%. The APH for pedestrians increases by 2.93%, 3.18%, and 2.42%, and the APH for cyclists increases by 2.84%, 3.21%, and 3.15%. These results demonstrate that our method achieves competitive performance in detecting small objects.

We also evaluate the methods across three distance ranges: 0-30m, 30-50m, and 50m-inf. The detection results for vehicles, pedestrians, and cyclists are shown in Tables 5, 6, and 7, respectively. For the results using Voxel R-CNN as the baseline, on level 1 difficulty, our method improves the APH for the vehicle class by 1.54%, 2.18%, and 2.39% across the three distance ranges, respectively. For the pedestrian class, our method improves the APH by 2.89%, 3.41%, and 3.22%, and for the cyclist class, our method improves the APH by 2.35%, 3.51%, and 4.84%. Our method shows more significant improvements in the 30-50m and 50m-inf distance ranges, which indicates that our method is superior in detecting distant objects. Detecting objects earlier allows for more time for subsequent behavior decisions and path planning, resulting in higher safety levels. Therefore, improving the detection performance of distant objects holds great application value.

| Difficulty | Methods | Veh. (3D mAP) (%) | | | | Veh. (3D mAPH) (%) | | | |
|---|---|---|---|---|---|---|---|---|---|
| | | Overall | 0-30m | 30-50m | 50m-Inf | Overall | 0-30m | 30-50m | 50m-Inf |
| LEVEL1 | PointRCNN [20] | 43.93 | 67.81 | 33.31 | 24.42 | 43.41 | 67.29 | 32.73 | 23.85 |
| | **Ours-P** | **45.65** | **69.28** | **35.34** | **26.71** | **45.19** | **68.78** | **34.84** | **26.18** |
| | Increasing value | 1.72 | 1.47 | 2.03 | 2.29 | 1.78 | 1.49 | 2.11 | 2.33 |
| | Voxel R-CNN [16] | 75.61 | 91.37 | 74.26 | 53.76 | 74.98 | 90.90 | 73.55 | 52.71 |
| | **Ours-V** | **77.47** | **92.85** | **76.37** | **56.13** | **76.92** | **92.44** | **75.73** | **55.10** |
| | Increasing value | 1.86 | 1.48 | 2.11 | 2.37 | 1.94 | 1.54 | 2.18 | 2.39 |
| | PV-RCNN [23] | 75.65 | 91.35 | 73.92 | 52.99 | 75.00 | 90.87 | 73.16 | 51.88 |
| | **Ours-PV** | **77.28** | **92.60** | **76.07** | **54.52** | **76.68** | **92.20** | **75.38** | **53.59** |
| | Increasing value | 1.63 | 1.25 | 2.15 | 1.53 | 1.68 | 1.33 | 2.22 | 1.71 |
| LEVEL2 | PointRCNN [20] | 37.91 | 66.51 | 29.41 | 18.00 | 37.46 | 66.00 | 28.89 | 17.58 |
| | **Ours-P** | **39.76** | **68.08** | **31.57** | **20.33** | **39.37** | **67.59** | **31.13** | **19.96** |
| | Increasing value | 1.85 | 1.57 | 2.16 | 2.33 | 1.91 | 1.59 | 2.24 | 2.38 |
| | Voxel R-CNN [16] | 66.82 | 90.09 | 67.61 | 41.59 | 66.26 | 89.62 | 66.94 | 40.74 |
| | **Ours-V** | **68.85** | **91.85** | **69.72** | **44.14** | **68.35** | **91.40** | **69.14** | **43.37** |
| | Increasing value | 2.03 | 1.76 | 2.11 | 2.55 | 2.09 | 1.78 | 2.20 | 2.63 |
| | PV-RCNN [23] | 67.08 | 90.04 | 67.27 | 40.92 | 66.48 | 89.57 | 66.56 | 40.03 |
| | **Ours-PV** | **68.70** | **91.35** | **69.48** | **42.56** | **68.17** | **90.93** | **68.84** | **41.79** |
| | Increasing value | 1.62 | 1.31 | 2.21 | 1.64 | 1.69 | 1.36 | 2.28 | 1.76 |

Table 5. Comparison on Waymo Open Dataset with 202 validation sequences for vehicle detection.

| Difficulty | Methods | Ped. (3D mAP) (%) | | | | Ped. (3D mAPH) (%) | | | |
|---|---|---|---|---|---|---|---|---|---|
| | | Overall | 0-30m | 30-50m | 50m-Inf | Overall | 0-30m | 30-50m | 50m-Inf |
| LEVEL1 | PointRCNN [20] | 18.93 | 25.97 | 13.51 | 11.49 | 16.96 | 23.62 | 11.75 | 9.75 |
| | **Ours-P** | **21.87** | **28.10** | **16.85** | **14.91** | **19.94** | **25.79** | **15.16** | **13.19** |
| | Increasing value | 2.94 | 2.13 | 3.34 | 3.42 | 2.98 | 2.17 | 3.41 | 3.44 |
| | Voxel R-CNN [16] | 74.34 | 79.95 | 73.39 | 62.40 | 63.75 | 70.78 | 61.56 | 49.01 |
| | **Ours-V** | **77.40** | **82.82** | **76.76** | **65.58** | **66.84** | **73.67** | **64.97** | **52.23** |
| | Increasing value | 3.06 | 2.87 | 3.37 | 3.18 | 3.09 | 2.89 | 3.41 | 3.22 |
| | PV-RCNN [23] | 72.78 | 77.66 | 72.34 | 61.85 | 61.35 | 67.69 | 59.45 | 47.48 |
| | **Ours-PV** | **75.13** | **79.40** | **75.12** | **64.19** | **63.64** | **69.38** | **62.24** | **49.89** |
| | Increasing value | 2.35 | 1.74 | 2.78 | 2.34 | 2.29 | 1.69 | 2.79 | 2.41 |
| LEVEL2 | PointRCNN [20] | 15.85 | 23.61 | 11.55 | 8.13 | 14.19 | 21.46 | 10.04 | 6.89 |
| | **Ours-P** | **18.74** | **25.98** | **14.81** | **11.44** | **17.12** | **23.84** | **13.37** | **10.28** |
| | Increasing value | 2.89 | 2.37 | 3.26 | 3.31 | 2.93 | 2.38 | 3.33 | 3.39 |
| | Voxel R-CNN [16] | 65.45 | 74.69 | 65.80 | 48.35 | 55.90 | 65.97 | 55.02 | 37.74 |
| | **Ours-V** | **68.58** | **77.44** | **69.26** | **51.62** | **59.08** | **68.76** | **68.53** | **40.82** |
| | Increasing value | 3.13 | 2.75 | 3.46 | 3.27 | 3.18 | 2.79 | 3.51 | 3.28 |
| | PV-RCNN [23] | 63.86 | 72.21 | 64.72 | 47.91 | 53.63 | 62.82 | 53.03 | 36.56 |
| | **Ours-PV** | **66.24** | **74.00** | **67.69** | **50.25** | **56.05** | **64.63** | **56.01** | **38.98** |
| | Increasing value | 2.38 | 1.79 | 2.97 | 2.34 | 2.42 | 1.81 | 2.98 | 2.42 |

Table 6. Comparison on Waymo Open Dataset with 202 validation sequences for pedestrian detection.

| Difficulty | Methods | Cyc. (3D mAP) (%) | | | | Cyc. (3D mAPH) (%) | | | |
|---|---|---|---|---|---|---|---|---|---|
| | | Overall | 0-30m | 30-50m | 50m-Inf | Overall | 0-30m | 30-50m | 50m-Inf |
| LEVEL1 | PointRCNN [20] | 45.43 | 55.72 | 35.54 | 36.44 | 43.96 | 54.04 | 33.77 | 35.09 |
| | **Ours-P** | **48.26** | **57.83** | **38.47** | **39.92** | **46.83** | **56.19** | **36.75** | **38.65** |
| | Increasing value | 2.83 | 2.11 | 2.93 | 3.48 | 2.87 | 2.15 | 2.98 | 3.56 |
| | Voxel R-CNN [16] | 69.88 | 81.32 | 64.94 | 49.32 | 68.27 | 79.94 | 63.20 | 46.58 |
| | **Ours-V** | **73.14** | **83.59** | **68.37** | **54.14** | **71.60** | **82.29** | **66.71** | **51.42** |
| | Increasing value | 3.26 | 2.27 | 3.43 | 4.82 | 3.33 | 2.35 | 3.51 | 4.84 |
| | PV-RCNN [23] | 66.80 | 78.60 | 61.81 | 46.71 | 64.89 | 76.91 | 60.11 | 42.29 |
| | **Ours-PV** | **69.94** | **81.26** | **64.93** | **49.95** | **68.08** | **79.58** | **63.30** | **45.61** |
| | Increasing value | 3.14 | 2.66 | 3.12 | 3.24 | 3.19 | 2.67 | 3.19 | 3.32 |
| LEVEL2 | PointRCNN [20] | 43.69 | 55.32 | 33.52 | 33.87 | 42.28 | 53.65 | 31.85 | 32.61 |
| | **Ours-P** | **46.50** | **57.30** | **36.79** | **37.46** | **45.12** | **55.61** | **35.19** | **36.22** |
| | Increasing value | 2.81 | 1.98 | 3.27 | 3.59 | 2.84 | 1.96 | 3.34 | 3.61 |
| | Voxel R-CNN [16] | 67.31 | 80.74 | 61.42 | 45.93 | 65.75 | 79.36 | 59.78 | 43.38 |
| | **Ours-V** | **70.46** | **82.82** | **64.59** | **50.56** | **68.96** | **81.57** | **63.14** | **47.99** |
| | Increasing value | 3.15 | 2.08 | 3.17 | 4.63 | 3.21 | 2.21 | 3.36 | 4.61 |
| | PV-RCNN [23] | 64.34 | 78.03 | 58.49 | 43.51 | 62.51 | 76.35 | 56.88 | 39.38 |
| | **Ours-PV** | **67.30** | **80.05** | **61.26** | **46.62** | **65.66** | **78.70** | **59.82** | **43.36** |
| | Increasing value | 2.96 | 2.02 | 2.77 | 3.11 | 3.15 | 2.35 | 2.97 | 3.98 |

Table 7. Comparison on Waymo Open Dataset with 202 validation sequences for cyclist detection.

**4.5 Ablation Study**

To analyze the effects of each proposed module, we conduct ablation studies using Voxel R-CNN as the baseline on the KITTI validation split. The results are listed in Table 8.

| Methods | TAFE | PSCL | Car (Mod.) (%) | Ped. (Mod.) | Cyc. (Mod.) |
|---------|------|------|----------------|-------------|-------------|
| A       |      |      | 84.52          | 60.67       | 73.06       |
| B       | √    |      | 84.87          | 61.62       | 73.93       |
| C       |      | √    | 84.76          | 62.03       | 73.86       |
| D       | √    | √    | **85.18**      | **63.18**   | **74.52**   |

Table 8. Effects of different modules on the moderate level of KITTI validation split. TAFE: Template-Assisted Feature Enhancement Module. PSCL: Proposal-Level Supervised Contrastive Learning Mechanism. The 3D detection results are evaluated with AP calculated by 11 recall positions.

**Effects of Template-Assisted Feature Enhancement Module**

Method A refers to the baseline method Voxel R-CNN, and Method B incorporates the template-assisted feature enhancement module. By utilizing templates with complete structure and dense points to provide intrinsic features, the network can learn richer structural information for foreground objects, effectively addressing the problems caused by uneven distribution and incomplete structures. For the car, pedestrian, and cyclist classes, the moderate AP is improved by 0.35%, 0.95%, and 0.87%, respectively, proving the effectiveness of the template-assisted feature enhancement module in 3D object detection.

**Effects of Proposal-level Supervised Contrast Learning Mechanism**

Method C incorporates the proposal-level supervised contrastive learning to Voxel R-CNN. For the three classes, the AP for moderate difficulty is improved by 0.24%, 1.36%, and 0.80%, respectively. Particularly, the improvement is most significant in the pedestrian class. The experimental results demonstrate that the contrastive learning not only further enhances the foreground features, but also suppresses the background features, thereby reducing false detections.

# 5. Conclusion

We propose an intrinsic feature guided 3D object detection method based on templates and contrast learning to address the issues caused by sparsity, incomplete structures and uneven distribution of point clouds. The proposed template-assisted feature enhancement module introduces templates with complete structures and dense points, and uses intrinsic features extracted from templates as optimization targets to guide the network in learning more discriminative foreground features. Additionally, to further distinguish between foreground and background objects, a proposal-level supervised contrastive learning mechanism is adopted to suppress background features. Experimental results on both the KITTI and Waymo Open datasets demonstrate that the two proposed modules can serve as plug-and-play components to embed in different base detectors, and improve their detection performance while maintaining the inference speed.

# Funding

This work was supported by the National Natural Science Foundation of China (NSFC, 61828501).

# Declaration of Competing Interest

The authors declare that they have no known competing financial interests or personal relationships that could have appeared to influence the work reported in this paper.

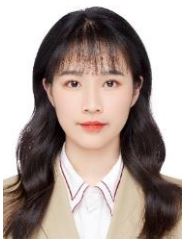

**Wanjing Zhang** received the bachelor's degree in electrical engineering and automation from Soochow University, China, in 2022.

she is currently pursuing the master degree with the School of Automation, Southeast University, Nanjing, China. Her research interests include 3D point cloud processing and 3D object detection.

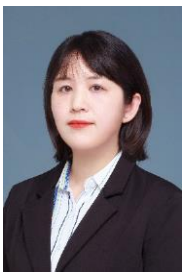

**Chenxing Wang** (Member, IEEE) received her Ph.D. degree from Southeast University in 2013, and was a research fellow at Nanyang Technological University in Singapore from 2014 to 2016.

She is now an associate professor at Southeast University. Her research interests include optical 3D measurement, optical computational imaging, 3D point cloud processing, human 3D reconstruction, and precision engineering.